\documentclass[10pt,twocolumn,letterpaper]{article}

\usepackage{cvpr}              

\usepackage{graphicx}
\usepackage{amsmath}
\usepackage{subcaption}
\usepackage{amsmath}
\usepackage{metawriting/metawriting}
\usepackage{multirow}
\usepackage{color}
\usepackage{amssymb}
\usepackage{subcaption}

\ours{REL-SF4PASS}
\ourHHA{REL representation}
\Eangle{the angle between surface normal and gravity direction}
\FIRSTH{horizontal disparity}
\SECONDH{height above ground}
\usepackage{metawriting/sciabbr}

\definecolor{cvprblue}{rgb}{0.21,0.49,0.74}
\usepackage[pagebackref,breaklinks,colorlinks,allcolors=cvprblue]{hyperref}
\def\paperID{30817} 
\def\confName{CVPR}
\def\confYear{2026}

\title{\ours: Panoramic Semantic Segmentation with REL Depth Representation and Spherical Fusion}

\author{Xuewei Li$^{1}$, Xinghan Bao$^{1}$, Zhimin Chen$^{1}$, Xi Li$^{2*}$ \\
$^{1}$School of Electronic and Information Engineering, Shanghai DianJi University\\
$^{2}$College of Computer Science and Technology, Zhejiang University \\
}

\begin{document}
\maketitle
\begin{abstract}
As an important and challenging problem in computer vision, Panoramic Semantic Segmentation (PASS) aims to give complete scene perception based on an ultra-wide angle of view. 
Most PASS methods often focus on spherical geometry with RGB input or using the depth information in original or HHA format, which does not make full use of panoramic image geometry. 
To address these shortcomings, we propose \textbf{\ours} with our \textbf{REL} depth representation based on cylindrical coordinate and \textbf{S}pherical-dynamic Multi-Modal \textbf{F}usion (SMMF). 
REL is made up of \textbf{R}ectified Depth, \textbf{E}levation-Gained Vertical Inclination Angle, and \textbf{L}ateral Orientation Angle, which fully represents 3D space in cylindrical coordinate style and the surface normal direction. 
SMMF aims to ensure the diversity of fusion for different panoramic image regions and reduce the breakage of cylinder side surface expansion in ERP projection, which uses different fusion strategies to match the different regions in panoramic images. 
Experimental results show that \textbf{\ours} considerably improves performance and robustness on popular benchmark, Stanford2D3D Panoramic datasets. 
It gains 2.35\% average mIoU improvement on all 3 folds and reduces the performance variance by approximately 70\% when facing 3D disturbance. 
\end{abstract}
\section{Introduction}
\label{sec:intro}
\begin{figure}[tb]
    \centering
        \begin{subfigure}{0.415\linewidth}
             \includegraphics[width=1\linewidth]{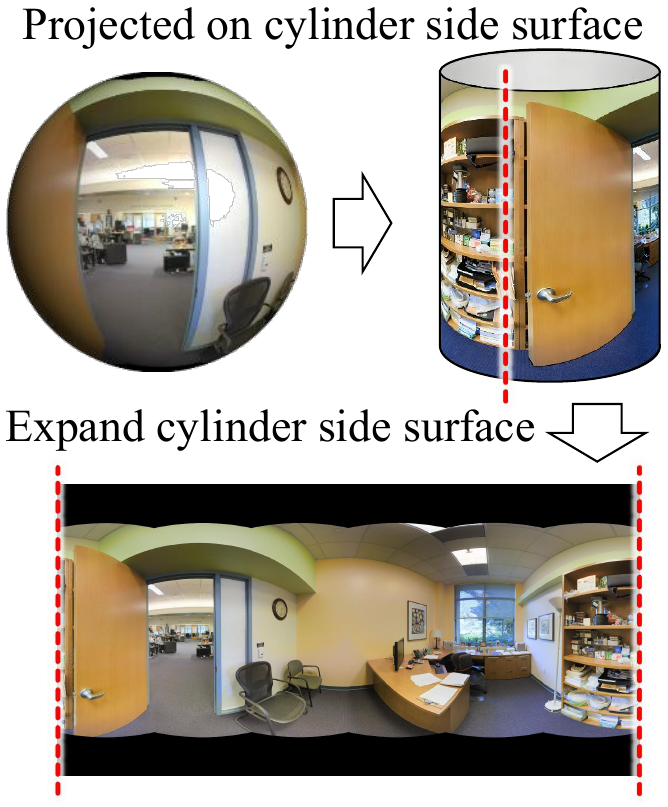}
            \caption{ERP projection. }
            \label{sfig:ERP}
        \end{subfigure} 
        \hspace{0.5cm}
        \begin{subfigure}{0.405\linewidth}
             \includegraphics[width=1\linewidth]{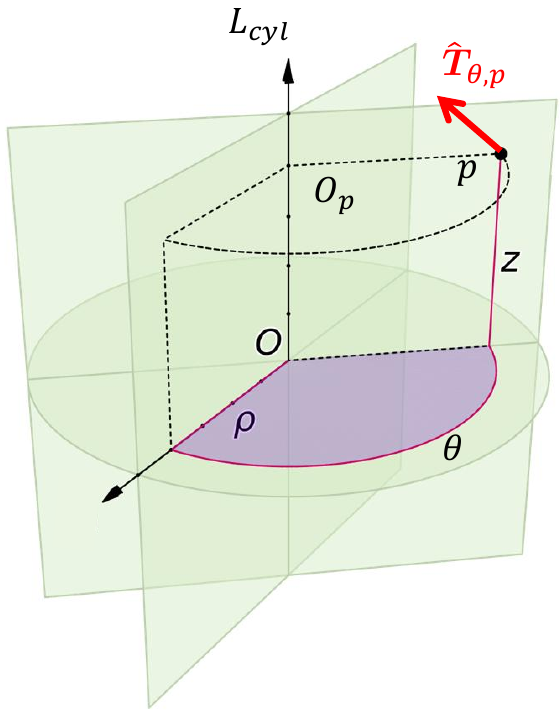}
            \caption{Cylindrical coordinate. }
            \label{sfig:CCS}
        \end{subfigure}

        \begin{subfigure}{1\linewidth}
            \includegraphics[width=1\linewidth]{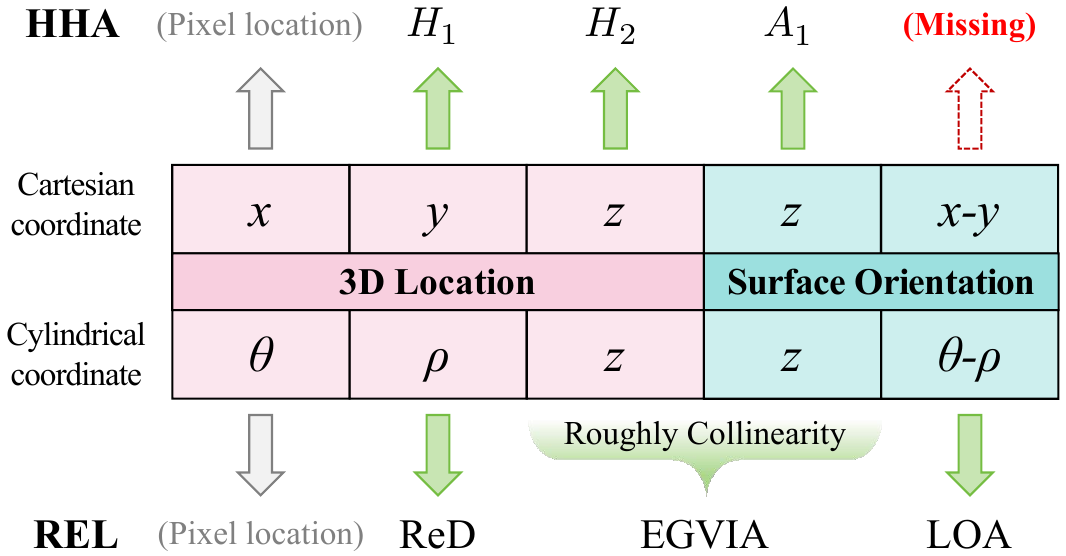}
            \caption{Comparison of HHA and REL. }
            \label{sfig:comapre}
        \end{subfigure}
    \caption{Overview of ERP projection, cylindrical coordinate and comparison of HHA and REL.
    $\hat{T}_{\theta, p}$ is the tangent line to circle $O_p$ at point $p$. 
    Circle $O_p$ lies in the plane perpendicular to gravity, passes through point $p$, and its center lies on the cylindrical axis $L_{cyl}$ of $\rho \theta z$. 
    In (c), we compared how different structure information are represented in REL and HHA.
    }
\end{figure}

There has been a great growing trend of practical applications based on $360^{\circ}$ cameras in recent years, including holistic sensing in autonomous vehicles~\cite{de2018eliminating,ma2021densepass,gao2022review}, immersive viewing in augmented reality and virtual reality devices~\cite{xu2018predicting, xu2021spherical, ai2022deep, zheng2023layoutdiffusion}, etc. 
Panoramic images with a complete Field of View (FoV, which is $360^{\circ} \times 180^{\circ}$) deliver complete scene perception in many real-world scenarios, thus drawing increasing attention in the research community in computer vision. 
Panoramic semantic segmentation (PASS) is essential for omnidirectional scene understanding, as it gives pixel-wise analysis for panoramic images and offers a dense prediction technical route acquiring complete perception of surrounding scenes~\cite{yang2021context}.
Most existing PASS approaches use equirectangular projection (ERP)~\cite{sun2021hohonet,yang2021capturing, feng2025spheredrag} to convert original $360^{\circ}$ data to 2D panoramic images. 
In brief, the original spherical information is first projected onto the side surface of a cylinder, and then the cylinder side surface is expanded as shown in~\cref{sfig:ERP}. 

Compared with RGB-only input, PASS strategies with RGB and depth input gain better performance, because depth information $D$ includes 3D location and shape (typically represented by surface normal) of each pixel beneficial to dense prediction like PASS.  
HHA~\cite{gupta2014learning} is widely used to explicitly model the 3D location and surface normal of each pixel in camera cartesian coordinate with \FIRSTH (first H, $H_1$), \SECONDH (second H, $H_2$), and \Eangle (last A, $A_1$), which gain considerable improvement compared with directly using original $D$. 
3D location has 3 degrees of freedom when surface normal direction has 2 degrees of freedom.  
As shown in~\cref{sfig:comapre}, HHA representation utilizes 3 different channels ($H_1$, $H_2$, and $A_1$) and the pixel location in the image to represent 4 degrees of freedom information.
However, the representation of the second degree of freedom for the surface normal direction is missing. 
Furthermore, the HHA calculation depends on the camera  posture and intrinsics (e.g., the focal length), which is not comfortable enough to deal with image-only data.

Facing this limitation and inspired by the cylinder used in ERP, we propose \ourHHA and directly use cylindrical coordinate $\rho \theta z$ to represent the 3D location and surface normal of each pixel.
\ourHHA is composed of \textbf{R}ectified Depth (ReD), \textbf{E}levation-Gained Vertical Inclination Angle (EGVIA), and \textbf{L}ateral Orientation Angle (LOA). 
Firstly, as shown in~\cref{sfig:comapre}, for the point $p$ in 3D space that corresponds to a certain pixel, we calculate the plane distance $\rho$, the plane angle $\theta$, and the height $z$ (the same as $H_2$) to directly represent the 3D location. 
Secondly, we calculate \Eangle and lateral orientation angle which is the angle between surface normal and tangent line $\hat{T}_{\theta, p}$. 
Also, \ourHHA does not need camera posture and intrinsics. 
Furthermore, we find roughly collinearity between $H_2$ and $A_1$ in panoramic images (discussed in~\cref{fig:HA-SEE} in detail), we fuse $H_2$ and $A_1$ as our EGVIA, set planar distance $\rho$ as our ReD, use lateral orientation angle as LOA, and gain our \ourHHA. 
It has 3 channels and is the same as HHA for a relatively fair comparison. 

In addition, due to the different local situations (e.g., distortion, location, etc.) of different regions in the panoramic image caused by ERP, we propose our Spherical-dynamic Multi-Modal Fusion (SMMF) to effectively fuse the RGB and REL information for different regions of panoramic images by different ways. 
Instead of achieving regions from input images, SMMF uses overlapping regions sampled on the cylinder side surface to reduce the breakage to the scene structure caused by cylinder side surface expansion.  
It not only allows different fusion ways for different regions, but also extracts semantic features across the left and right edges of a panoramic images.
Using \ourHHA and SMMF jointly, we build up \ours, and our contributions are summarized as follows:
\begin{itemize}
    \item We propose our 3-channel \ourHHA to represent the 3D location and shape of each pixel based on cylindrical coordinate.  
    \item Our Spherical-dynamic Multi-Modal Fusion (SMMF) enriches the diversity of fusion for different regions in panoramic images and reduce the impact of cylinder side surface expansion by ERP projection.
    \item We evaluate \ours on popular Stanford2D3D Panoramic datasets, and gain the 2.35\% average mIoU improvement on all 3 folds. 
    Also, it reduces the performance variance by about 70\% in SGA validation.     
\end{itemize}

\section{Related Work}
\label{sec:related}
The two most related fields are panoramic semantic segmentation and RGB-D semantic segmentation. 

\begin{figure*}[tb]
    \centering
            \includegraphics[width=\linewidth]{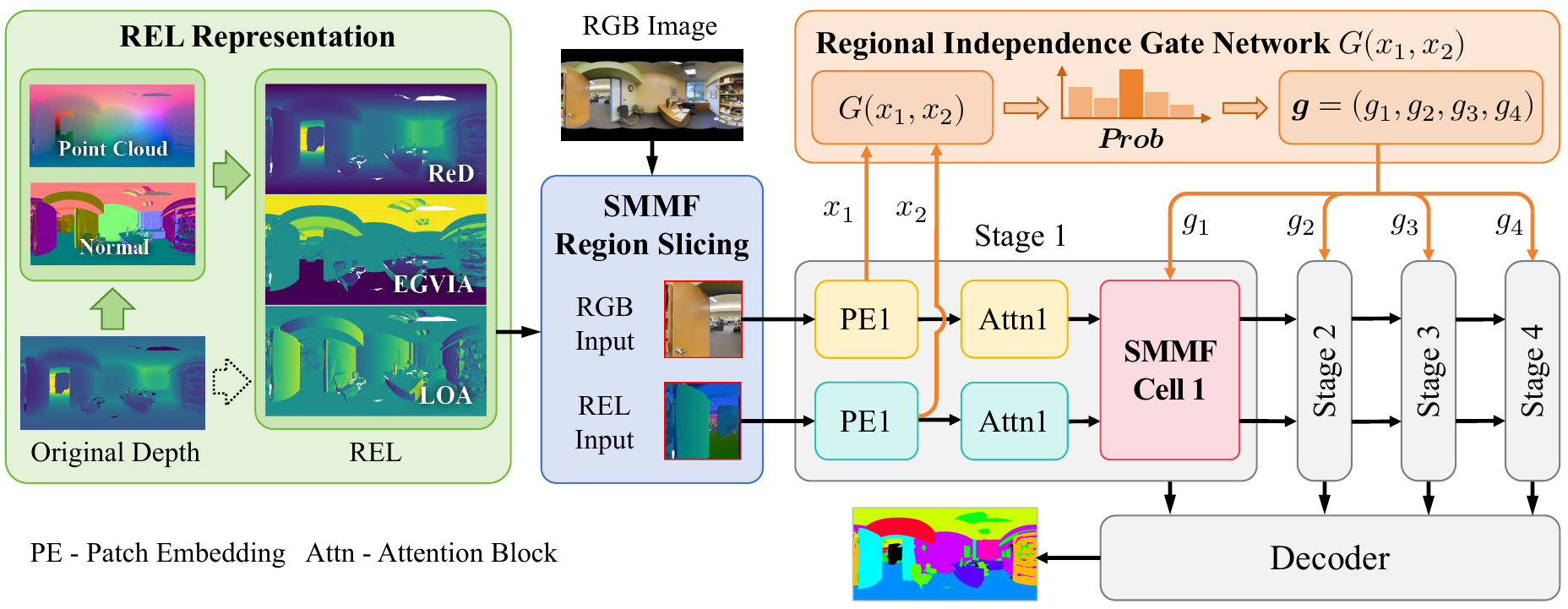}
            \caption{
           The overview of \ours.
           Firstly, \ourHHA effectively represents the depth information by using ReD, EGVIA, and LOA, which contains the 3D location and surface normal direction.
           Secondly, Our SMMF to get image regions from cylinder side surface and uses a Gate Network to independently determine the fusion used in each region.
            }
            \label{fig:main}
\end{figure*}

\subsection{Panoramic Semantic Segmentation}
\label{ssec:PASS}
PAnoramic Semantic Segmentation (PASS) is the task that does semantic segmentation on panoramic images. 
In recent years, many models have been developed for PASS.
Deng et al.~\cite{deng2017cnn} built up the first semantic segmentation framework for big FoV (wide-angle / fish-eye) images. 
Furthermore, Yang et al.~\cite{yang2020ds} used attention connections to make PASS more efficient, and DS-PASS was proposed. 
Common PASS solutions can be divided into two main fields: geometry-aware strategies and distortion-aware ones. 
For geometry-aware strategies~\cite{pintore2021slicenet, li2023sgat4pass, liu2025360, liu2025twin}, some utilized 2D geometry like horizontal features mainly based on the ERP inherent property when some use 3D geometry. 
Tateno et al.~\cite{tateno2018distortion} used carefully designed distortion-aware convolutions to deal with image distortions when Jiang et al.~\cite{jiang2019spherical} utilized a spherical convolution operation. 
SliceNet~\cite{pintore2021slicenet} and HoHoNet~\cite{sun2021hohonet} used 1D horizontal representation to construct their feature maps.
ACDNet~\cite{zhuang2022acdnet} proposed convolution equipped with different dilation rates adaptively in PASS. 
Trans4PASS~\cite{zhang2022bending} and Trans4PASS+~\cite{zhang2024behind} proposed their Deformable Patch Embedding (DPE) and Deformable Multi-Layer Perception (DMLP) modules to perceive and deal with spherical distortion.
SGAT4PASS~\cite{li2023sgat4pass} designed a reprojection strategy to augment training samples and used SDPE to apply the spherical symmetry into the deformable components in the model.
For distortion-aware strategies, Lee et al.~\cite{lee2018spherephd} represented panoramic views with spherical polyhedrons to minimize the difference in spatial resolution of the surface of the sphere.
Liu et al.~\cite{liu2025360} proposed their spherical discrete sampling based on the weights of the pre-trained models in order to mitigate distortions effectively. 


\subsection{RGB-D Semantic Segmentation}
\label{ssec:RGB-DSS}
RGB-D semantic segmentation takes advantage of both RGB and depth information and gets considerable performance improvements. 
Previous works mainly focus on two aspects: (a) new operators / layers / modules based on the geometric properties of RGB-D data~\cite{cao2021shapeconv, chen2021spatial, wang2018depth, xing2020malleable, hu2019acnet}; (b) making specialized architectures for jointly utilizing RGB and depth features~\cite{sun2021hohonet, park2017rdfnet, zhang2023cmx, zhang2019hyperfusion, guttikonda2024single}. 
For the former ones, models are carefully designed. 
E.g., ACNet~\cite{hu2019acnet} gained informative features from RGB-D data by attention. 
For the latter one, various RGB and depth information fusion ways have been explored.
E.g., CMX~\cite{zhang2023cmx} built CM-FRM to use one modality features to rectify the features of the other one, and proposed FFM to make enough long-range context interaction.

\ours focuses on RGB-D PASS task, we carefully design our \ourHHA for the geometric structure of panoramic images, and our SMMF as a spherical-guided region-adaptive dynamic multi-modal fusion.


\begin{figure*}[tb]
    \centering
        \includegraphics[width=1\linewidth]{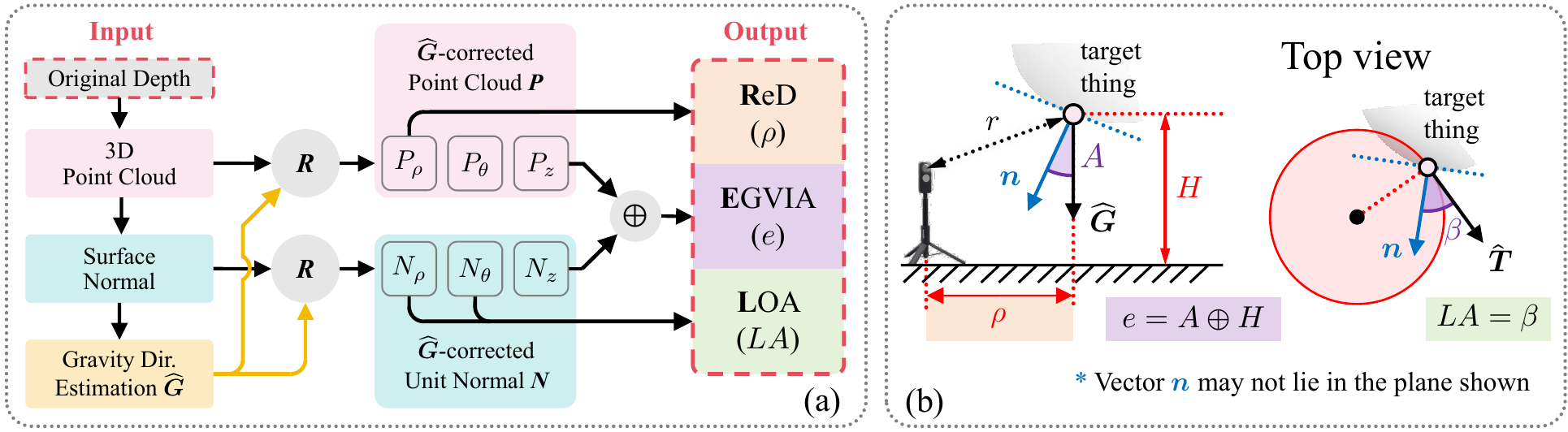}
        \caption{
             The calculation process (a) and physical model (b) of \ourHHA.
             In (a), $\mathbf{R}$ is 3D Rigid Body Rotation, and $\oplus$ is~\cref{eq:Ecal}. 
            }
            \label{fig:REL-detail}
\end{figure*}

\section{Method}
\label{method}
\ours is introduced in this section. 
First, we introduce the background, notations, and classic HHA in~\Cref{ssec:background}. 
Second, we propose our REL representation made up of Rectified Depth, Elevation-Gained Vertical Inclination Angle, and Lateral Orientation Angle. 
Furthermore, our Spherical-dynamic Multi-Modal Fusion (SMMF) utilizes depth information more efficiently  and is introduced in~\Cref{sssec:RDMM}.
The overall pipeline is shown in~\cref{fig:main}. 

\subsection{Background and Notations}
\label{ssec:background}
We first describe the common formulation of panoramic images and classic HHA. 
For expression convenience, we build up several coordinates: 

1) Image coordinate $uvd$ ($u$: pixel column (positive to the right), $v$: pixel row (positive downward), $d$: Pixel depth), and the origin is image upper left corner); 

2) Gravity-correction camera cartesian coordinate $xyz$ (right-handed): using camera as origin, first define $y$ positive direction as from camera to target pixel, and then gravity correction is performed with rectifying the direction of gravity as the negative $z$ direction.  

3) Spherical coordinate $r \theta \phi$ ($r$ is the radius, $\theta$ is the azimuth / longitude (from $-180^{\circ}$ to $180^{\circ}$), $\phi$ is the elevation / latitude (from $-90^{\circ}$ to $90^{\circ}$), and the origin is the camera).

\textbf{ERP projection} is widely-used to transform original $360^{\circ}$ data into 2D panoramic images. 
As shown in~\cref{sfig:ERP}, $360^{\circ}$ data is first projected onto the cylinder side surface and then expand the cylinder side surface. 
Original $360^{\circ}$ data is formulated in the spherical coordinate. 
To convert it to a rectangular image in the image coordinate, let $u = (\theta - \theta_0) cos \phi_1$ and $v = (\phi-\phi_1)$. 
$\theta_0 = 0$ is the central latitude and $\phi_1 = 0$ is the central longitude. 
The ERP-processed rectangular images are the input in PASS, and the rectangular semantic segmentation results are the final output to calculate metrics. 

\textbf{HHA representation} changes original depth information $D$ to 3 channels with the help of camera intrinsics: \FIRSTH (first H, $H_1$), \SECONDH (second H, $H_2$), and \Eangle (last A, $A_1$). 
It is calculated as (all values are linearly scaled to $[0, 255]$):
(1) Calculate $H_1$ from the original depth information $D$ and camera intrinsics (see Section A ``Details for Constructing 3D Point Cloud'' in the supplementary material);
(2) Convert $D$ to a 3D point cloud with camera intrinsics, gain the normal field, estimate gravity direction, calculate the gravity-corrected 3D point cloud $P$ and normal field $N$. 
(3) Calculate $H_2$ and $A_1$ using~\cite{gupta2013perceptual}. 
 
As shown in~\cref{sfig:comapre}, the representation of the second degree of freedom for the surface normal direction is missing.

    
\begin{figure*}[tb]
    \centering
        \includegraphics[width=1\linewidth]{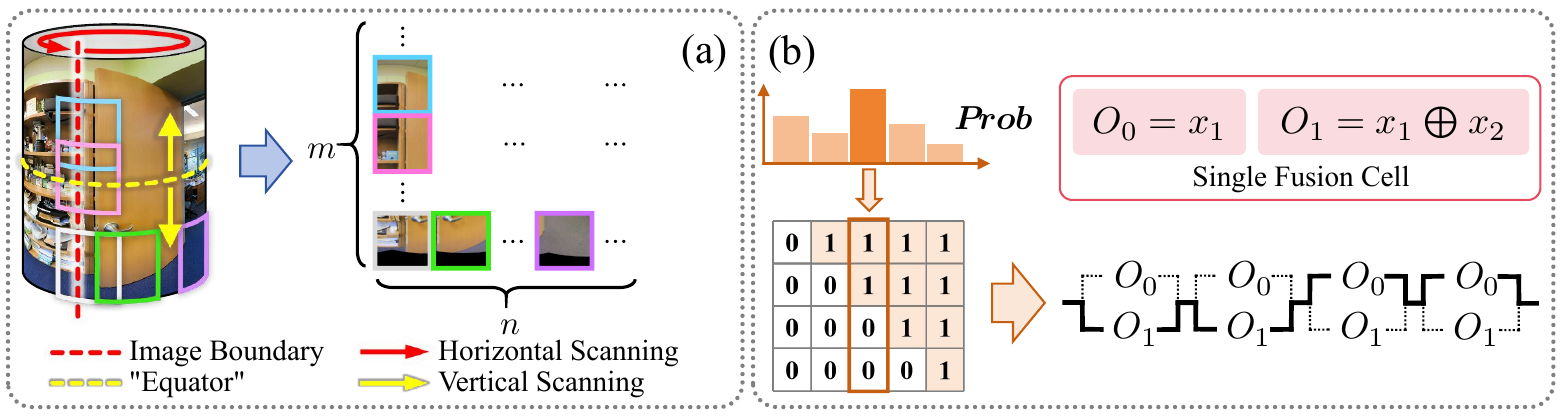}
        \caption{
             The detail design of SMMF Region Slicing (a) and Gate Network (b). 
             Slicings have overlap and $\oplus$ means the fusion operation. 
            }
            \label{fig:SMMF-detail}
\end{figure*}

\subsection{REL Representation}
\label{ssec:REL}
Dealing with the limitation above, we make full use of the geometry of panoramic images to design our REL directly represent the 3D location and surface normal of each pixel based on cylindrical coordinate $\rho \theta z$. 

Firstly, compared with plane images, panoramic images provide a complete FoV and carry rich prior information, such as using pixel location in image coordinate can derive the direction of corresponding point from camera without using camera intrinsics (see Section A ``Details for Constructing 3D Point Cloud'' in the supplementary material). 

Secondly, for a certain point $p$, its location in gravity-correction camera cartesian coordinate is $(p_x, p_y, p_z)$ and its unit surface normal is $(N^p_x, N^p_y, N^p_z)$. 
Our REL representation model the position in 3D space by planar distance, plane angle and height based on cylindrical coordinate $\rho \theta z$, and model unit normal $N$ with 2 independent angle, \Eangle and the angle between surface normal and planar tangent line $\hat{T}_{\theta, p}$. 
Our \textbf{R}ectified Depth (ReD) $\rho_p$ corresponds to the planar distance. 
The plane angle $\theta$ can be directly gained by pixel spherical coordinate $r \theta \phi$ representation. 
Our \textbf{E}levation-Gained Vertical Inclination Angle (EGVIA) $\hat{e_p}$ integrates 2 roughly collinearity components, height $H_p$ and \Eangle (indicating $N^p_z$) with normalization. 
Our \textbf{L}ateral Orientation Angle (LOA) indicates the angle between surface normal and planar tangent line $\hat{T}_{\theta, p}$. 
For $N$ is unit vector with 2 degrees of freedom, 2 angles are enough. 
The physical models of all 3 above are shown in the right part of~\cref{fig:REL-detail}. 

In calculation, as shown in the left part of~\cref{fig:REL-detail}, REL only needs original depth image input (without camera posture or intrinsics) to gain gravity-corrected 3D point cloud $P$ and achieve the normal field $N$, and all channels are calculated based on $P$ and $N$. 

\subsubsection{Rectified Depth}
\label{sssec:ReD} 
Rectified depth $\rho_p$ aims to represent the planar distance between camera and $p$, so it is defined as:
\begin{equation}
\label{eq:Rcal}
    \rho_p = \sqrt{p_x^2+p_y^2} = d_p \cos \phi_p,
\end{equation}
where $d_p$ is the depth distance of $p$, and $\phi_p$ is gained from the input image and gravity correction. 

\subsubsection{Elevation-Gained Vertical Inclination Angle}
\label{sssec:EGVIA}
To calculate elevation-gained vertical inclination angle of $p$, we should first gain the height $H_p$ (the same definition as $H_2$ in HHA) and \Eangle $A_p$ (the same as $A_1$ in HHA) of $p$. 
 we calculate the height $H_p$ as:
\begin{equation}
\label{eq:Heightcal}
    H_p = p_z - \min \{P_z\}, 
\end{equation}
where $\min \{P_z\}$ means the minimum value of $p_z$ among all $p$ in $P$. 
We calculate $A_p$ as: 
\begin{equation}
\label{eq:Acal}
    A_p = \arccos{(N^p \odot \mathbf{\hat{G}})} = \arccos{N^p_z}, 
\end{equation}
where $\odot$ is dot product, $N^p$ is the surface normal of $p$, $\mathbf{\hat{G}}$ is the unit vector of the gravity direction that is $(0, 0, -1)$ in $xyz$, and $N^p_z$ is the z-axis component of $N^p$.

\begin{figure}[tb]
    \centering
        \includegraphics[width=1\linewidth]{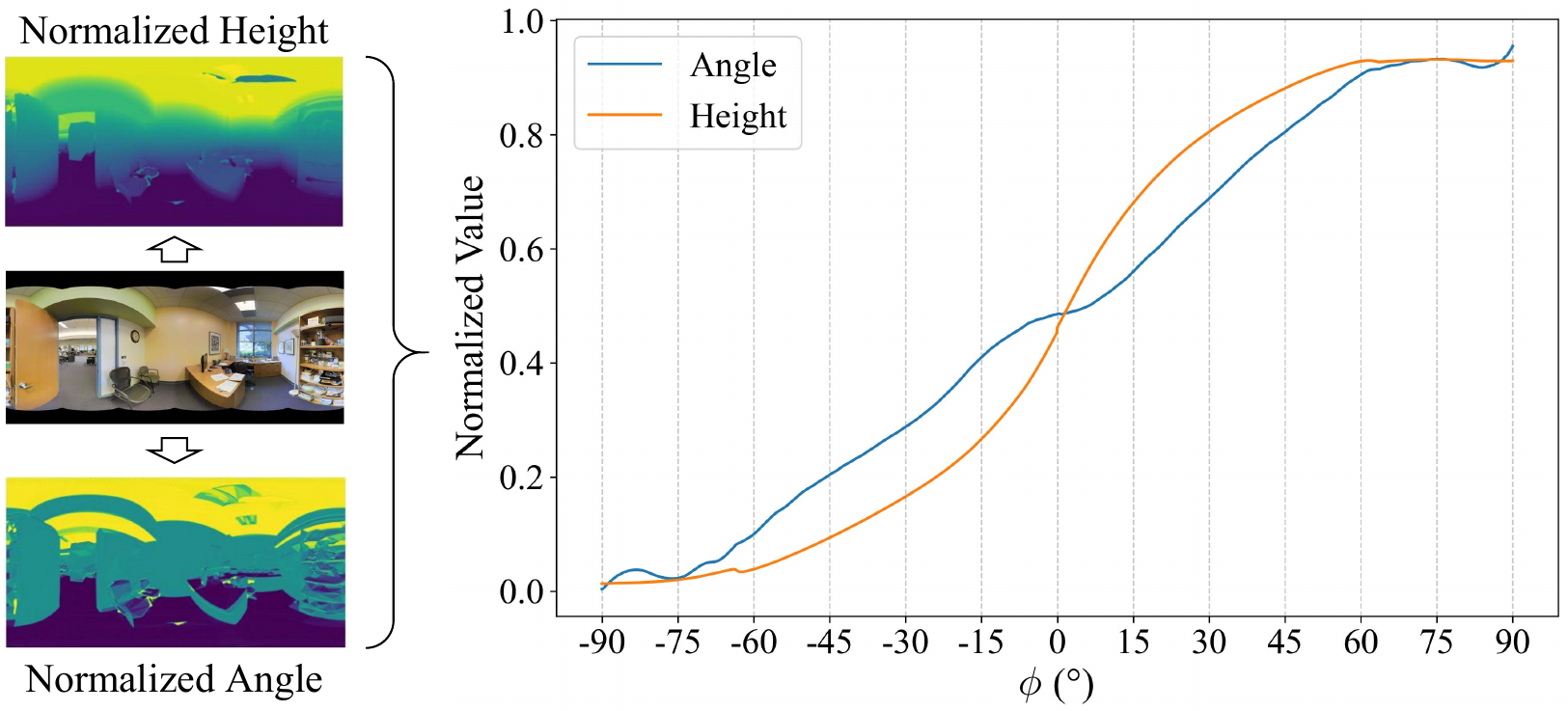}
        \caption{Comparison of normalized height and \Eangle (short as Angle). 
        The x-axis of the right figure is $\phi$ (from $-90^{\circ}$ to $90^{\circ}$) when the y one is the average of all normalized corresponding values.}
    \label{fig:HA-SEE}
    \centering
\end{figure}

For panoramic images with complete FoV, the change of camera direction does not affect the whole imaging content when its representation is mainly influenced by ERP projection. 
For humans, ERP projection has a popular and conventional usage, that is, the normal vector of bottom plane in the cylinder used for ERP projection is parallel to the direction of gravity. 
As a result, for panoramic images with common ERP projection, it is most the ground when $\phi$ is close to $-90^{\circ}$ and it is most the sky or ceiling when $\phi$ is close to $90^{\circ}$, which makes normalized $H_p$ and $A_p$ highly related. 
\cref{fig:HA-SEE} is a visualization.  
The main difference between $H_p$ and $A_p$ is: 
(1) For nearly horizontal surface (such as ground, tabletop, etc.), $H_p$ can distinguish different surface more directly than $A_p$; 
(2) For nearly vertical surface (such as wall, etc.), whose $A_p$ are close to $90^{\circ}$. 
Based on these observations and to represent this information more efficiently, we propose elevation-gained vertical inclination angle $\hat{e_p}$ of a certain point $p$ as: 
\begin{equation}
\label{eq:Ecal}
    \hat{e_p} = 
    \begin{cases}
        \lambda \hat{A_p} + (1 - \lambda) \hat{H_p}, & A_p \in [0^{\circ}, \alpha) \cup \\
         & \quad (180^{\circ}-\alpha, 180^{\circ}] \\
        \hat{A_p}, & A_p \in [\alpha, 180^{\circ}-\alpha]
	\end{cases} ,\\
\end{equation}

where $\hat{A_p}$ and $\hat{H_p}$ are normalized $A_p$ and $H_p$ (linearly scaled to $[0, 255]$~\cite{gupta2013perceptual}). 
Using elevation-gained vertical inclination angle $E_p$, we jointly represent height and \Eangle.





\subsubsection{Lateral Orientation Angle}
\label{sssec:LOA}
The rest component is the angle between surface normal and planar tangent line $\hat{T}_{\theta, p}$. 
In detail, $\hat{T}_{\theta, p}$ is the tangent line to circle $O_p$ at point $p$, where circle $O_p$ lies in the plane perpendicular to the direction of gravity, its center lies on the cylindrical axis $L_{cyl}$ of $\rho \theta z$, and passes through point $p$. 
we define our lateral orientation angle $LA_p$ to represent it of $p$ as:
\begin{align}
\label{eq:Lcal}
    \hat{LA_p} & = \mathrm{Norm}(\arccos{(N^p \odot \mathbf{\hat{T}})}) \\
        & = \mathrm{Norm}(\arccos{(N^p_x \cdot \cos \theta + N^p_y \cdot \sin \theta)}), 
\end{align}
where $\odot$ is dot product, $\mathbf{\hat{T}}$ is the unit vector of tangent to the latitude that $p$ belongs to, $(\cos \theta, \sin \theta, 0)$, and $N^p_x$ / $N^p_y$ is the x-axis / y-axis component of $N^p$. 
$\mathrm{Norm}(\cdot)$ is the normalization for linearly scaled to $[0, 255]$~\cite{gupta2013perceptual}.



\begin{table}[]
    \footnotesize
    \centering
    \resizebox{0.42\textwidth}{!}{
    \begin{tabular}{l|c|c|c}
    \hline
    Method                                     & modal   & Avg mIoU & F1 mIoU \\
    \hline
    StdConv (2018)~\cite{tateno2018distortion}        &  RGB     &  -    & 32.6  \\
    CubeMap (2018)~\cite{tateno2018distortion}        &  RGB     &  -    & 33.8  \\
    DistConv (2018)~\cite{tateno2018distortion}       &  RGB     &  -    & 34.6  \\	
    SWSCNN (2020)~\cite{esteves2020spin}              &  RGB     & 43.4  &  -    \\
    Tangent (2020)~\cite{tangent}                     &  RGB     & 45.6  &  -    \\
    HoHoNet (2021)~\cite{sun2021hohonet}              &  RGB     & 52.0  & 53.9 \\
    PanoFormer (2022)~\cite{shen2022panoformer}       &  RGB     & 48.9  & -    \\
    Trans4PASS (2022)~\cite{zhang2022bending}         &  RGB     & 52.1  & 53.3 \\
    FreDSNet (2023)~\cite{berenguel2023fredsnet}      &  RGB     & -     & 46.1  \\
    CBFC (2023)~\cite{zheng2023complementary}         &  RGB     & 52.2  & -    \\
    SGTA4PASS (2023)~\cite{li2023sgat4pass}           &  RGB     & 55.3  & 56.4 \\
    Trans4PASS+ (2024)~\cite{zhang2024behind}         &  RGB     & 53.7  & 53.6 \\
    Twin (2025)\cite{liu2025twin}                     &  RGB     & 55.85 & - \\
    \hline
    Tangent (2020)~\cite{tangent}                     &  RGB-D   & 52.50 & - \\
    HoHoNet (2021)~\cite{sun2021hohonet}              &  RGB-D   & 56.73 & - \\
    PanoFormer (2022)~\cite{shen2022panoformer}       &  RGB-D   & 57.03 & - \\
    CBFC (2023)~\cite{zheng2023complementary}         &  RGB-D   & 56.70 & - \\
    SFSS (2024)~\cite{guttikonda2024single}           &  RGB-D   & 55.49 & - \\
    \hline
    SFSS (2024)~\cite{guttikonda2024single}           &  RGB-N   & 59.43 & - \\
    SFSS (2024)~\cite{guttikonda2024single}           &  RGB-HHA-D & 59.99 & - \\
    SFSS (2024)~\cite{guttikonda2024single}           &  RGB-HHA-N & 60.24 & - \\
    SFSS (2024)~\cite{guttikonda2024single}           &  RGB-HHA & 60.60 & - \\
    CMX*~\cite{zhang2023cmx}                     &  RGB-HHA & 60.71 & 63.98 \\
    \hline
    Ours                                              &  RGB-REL & \textbf{63.06} & \textbf{67.37} \\
    \end{tabular}
    }
    \caption{Comparison with the SOTA methods on Stanford2D3D Panoramic datasets with traditional metrics. 
    We follow recent works to compare the performance of both official fold 1 and the average performance of all three official folds. respectively. 
    ``Avg mIoU'' / ``F1 mIoU'' means the mIoU performance of three official folds on average / official fold 1. 
    The performance from StdConv (2018) to Trans4PASS+ (2024) are directly cited from~\cite{li2023sgat4pass}.
    Following, the performance from Twin (2025) to SFSS (2024) with RGB-D are directly cited from \cite{liu2025twin} and no fold 1 mIoU is provided. 
    SFSS (2024)~\cite{guttikonda2024single} from RGB-N to RGB-HHA are directly cited from~\cite{guttikonda2024single} and ``-N'' means using normals modalities. 
    CMX* is our implementation for CMX~\cite{zhang2023cmx} on this PASS task. 
    Considerable improvement is gained and our \ourHHA is more suitable for model training.  
    }
    \label{tab:sota}
\end{table}

\renewcommand{\arraystretch}{1.5}
\begin{table*}
    \footnotesize
    \centering
    \resizebox{0.99\textwidth}{!}{
        \begin{tabular}{c|c|c|c|c|c|c|c}
        \toprule
        \multirow{2}{*}{($\beta$,$\gamma$,$\alpha$) ($^{\circ}$)} & HHA mIoU / PAcc   & \multirow{2}{*}{($\beta$,$\gamma$,$\alpha$) ($^{\circ}$)} & HHA mIoU / PAcc   & \multirow{2}{*}{($\beta$,$\gamma$,$\alpha$) ($^{\circ}$)} & HHA mIoU / PAcc   & \multirow{2}{*}{($\beta$,$\gamma$,$\alpha$) ($^{\circ}$)} & HHA mIoU / PAcc   \\ \cline{2-2} \cline{4-4} \cline{6-6} \cline{8-8} 
                                                 & Our mIoU / PAcc &                                              & Our mIoU / PAcc &                                              & Our mIoU / PAcc &                                              & Our mIoU / PAcc \\ \midrule
        \multirow{2}{*}{(0,0,0)}                     & 65.433 / 90.786        & \multirow{2}{*}{(0,5,0)}                     & 61.877 / 88.571	         & \multirow{2}{*}{(5,0,0)}                   & 61.252 / 88.310	           & \multirow{2}{*}{(5,5,0)}                     & 59.340 / 87.155         \\ \cline{2-2} \cline{4-4} \cline{6-6} \cline{8-8} 
                                                    & 67.367 / 90.912		  &                                              & 65.391 / 89.846		     &                                            & 65.251 / 89.730		           &                                              & 63.835 / 89.048         \\ \hline
        \multirow{2}{*}{(0,0,90)}                    & 65.597 / 90.745          & \multirow{2}{*}{(0,5,90)}                    & 61.525 / 88.371         & \multirow{2}{*}{(5,0,90)}                  & 62.515 / 88.880           & \multirow{2}{*}{(5,5,90)}                    & 59.130 / 86.976         \\ \cline{2-2} \cline{4-4} \cline{6-6} \cline{8-8} 
                                                    & 67.010 / 90.834          &                                              & 65.138 / 89.755	         &                                            & 65.285 / 89.799         &                                              & 63.595 / 88.850         \\ \hline
        \multirow{2}{*}{(0,0,180)}                   & 65.507 / 90.764         & \multirow{2}{*}{(0,5,180)}                   & 62.363 / 88.924         & \multirow{2}{*}{(5,0,180)}                  & 61.742 / 88.363         & \multirow{2}{*}{(5,5,180)}                   & 59.523 / 87.439         \\ \cline{2-2} \cline{4-4} \cline{6-6} \cline{8-8} 
                                                    & 67.223 / 90.891         &                                              & 65.490 / 89.848         &                                              & 65.475 / 89.959         &                                              & 64.286 / 89.282          \\ \hline
        \multirow{2}{*}{(0,0,270)}                   & 65.852 / 90.782         & \multirow{2}{*}{(0,5,270)}                   & 62.050 / 88.453         & \multirow{2}{*}{(5,0,270)}                  & 62.503 / 88.596         & \multirow{2}{*}{(5,5,270)}                   & 59.588 / 87.133         \\ \cline{2-2} \cline{4-4} \cline{6-6} \cline{8-8} 
                                                    & 67.145 / 90.904         &                                              & 65.179 / 89.877        &                                               & 65.019 / 89.780         &                                              & 63.393 / 88.948         \\ \bottomrule
        \end{tabular}
    }
    \caption{
    Detail performance comparison of REL and HHA both with SMMF on Stanford2D3D Panoramic datasets fold 1 with SGA metrics. 
    All 16 test situations are shown, and the analysis is in~\cref{tab:STA_SGAM_BIG}.  
    ``PAcc'' meas the pixel accuracy metric. 
    }
    \label{tab:SGAMBIG}
\end{table*}

\subsection{Spherical-dynamic Multi-Modal Fusion}
\label{sssec:RDMM}
After using REL to represent depth information, we need to fuse it and RGB information effectively. 
A common way is to fuse the two information after each block in the model but lacks diversity.
Sample-adaptive fusion~\cite{xue2023dynamic}, is classic but effective way for this problem. 
For panoramic images based on ERP projection, there is some structural prior. 
E.g., the pixels have the same latitude in the spherical coordinate so that their distortion and semantics are somehow similar. 
For example, the upper part (Latitude $\phi$ is close to $90^{\circ}$) of an indoor panoramic image is mostly the ceiling, the lower one (Latitude $\phi$ is close to $-90^{\circ}$) is mostly the floor, and the middle one (Latitude $\phi$ is close to $0^{\circ}$) usually has richer semantics.
It is necessary to adopt different fusion strategies for different latitude regions, that is, to refine the adaptive granularity of fusion from the sample level to the regional level.
Also, the cylinder side surface expansion in ERP projection breaks the scene structure in panoramic images.  

Inspired by DynMM~\cite{xue2023dynamic} and faced with 2 challenges above, we propose our Spherical-dynamic Multi-Modal Fusion (SMMF), which uses different fusion strategies for not only different inference images but also different regions in one image. 
Instead of achieving regions from panoramic images, SMMF uses overlapping regions directly sampled on the cylinder side surface to reduce the breakage to the scene structure caused by cylinder side surface expansion.
As shown in the left part of~\cref{fig:SMMF-detail}, we sample $m \times n$ (m rows and n columns) overlap regions on the cylinder side surface. 
In the horizontal direction, we allow the regions selecting across the left and right edges of the panoramic image to reduce the influence of cylindrical side surface expansion in ERP projection. 
In the vertical direction, different from common image region generation~\cite{xue2023dynamic}, we start from the ``equator'' ($\phi = 0^{\circ}$) and proceed sequentially towards the top and bottom edges of the image to ensure the obtained regions symmetrical about the equator. (E.g., begin from the pick square to the blue / gray square in the left part of~\cref{fig:SMMF-detail})
We take into account all related region fusion modes for the overlap area, and sum them as the final result. 

For the detailed RGB-D fusion design, we insert several fusion cell $FusCell_i$ to fuse 2 modalities, $M = \{M_{RGB}, M_{D}\}$.  
As widely-used Mixture-of-Experts (MoE) ways, for the fusion cell $FusCell_i$ in stage $i$, we make a set of $B$ expert fusion operations in parallel as $\{FusOp_i\}$, which specializes in a subset of two modalities. 
In particular, $M_{D}$ includes scene structure cues but often performs poorly by themselves in semantic segmentation without RGB input, so that $M_{D}$ is not used alone. 
A gate network $G(x)$ is designed to decide which expert network should be activated. 
It produces a $B$-dimensional vector $\mathbf{Prob} = [Prob_1, Prob_2, \cdots, Prob_B]$ for each $FusCell_i$. 
The output $OutCell_i$ of cell $FusCell_i$ is $OutCell_i = \sum^B_{i=1} Prob_i \cdot FusOp_i(x)$ with multi-modal input as shown in the right part of~\cref{fig:SMMF-detail}.  
Similar to DynMM~\cite{xue2023dynamic}, we use two-stage soft-hard training and early stop. 
The whole training is divided into two stages: soft training and hard training. 
For soft training stage, $\mathbf{Prob}$ is a ``soft'' probability vector whose each value may be non-zero when $\mathbf{Prob}$ must be one-hot in hard training stage. 
we set 0 means no fusion (using RGB only) and 1 means using RGB-D fusion for convenience. 
For fusion early stop, it means that if 0 (no fusion) is selected in a previous cell $FusCell_i$, all the following cells $FusCell_j, (j>i)$ are forced to 0.

Using \ourHHA and SMMF jointly, we clearly represent depth information and use a spherical adaptive method to effectively fuse multi-modal information.

\begin{table}
    \footnotesize
    \centering
    \resizebox{0.42\textwidth}{!}{
        \begin{tabular}{c|cccc}
            \hline
            \multirow{2}{*}{Statistics} & \multicolumn{2}{c}{HHA} & \multicolumn{2}{c}{\ours} \\
                                        & mIoU        & PAcc      & mIoU      & PAcc    \\
            \hline
            Mean                        & 62.237 	  & 88.766        & 65.380 ({\color{red}{+3.143}})    & 89.891 ({\color{red}{+1.126}}) \\
            Variance                    & 5.316 	  & 1.801         & 1.607 ({\color{red}{-3.709}})     & 0.472 ({\color{red}{-1.328}})  \\
            Range                       & 6.722       & 3.810         & 3.974 ({\color{red}{-2.748}})     & 2.062 ({\color{red}{-1.748}})  \\ 
            \hline
        \end{tabular}
    }
    \caption{
    Overall performance comparison with HHA with Fold 1 on Stanford2D3D Panoramic datasets based on \cref{tab:SGAMBIG}. 
    ``PAcc'' means the pixel accuracy metric. 
    \ourHHA earns considerable mean performance and robustness improvement.  
    }
    \label{tab:STA_SGAM_BIG}
\end{table}

\begin{table}
    \footnotesize
    \centering
    \resizebox{0.42\textwidth}{!}{
        \begin{tabular}{ccccc}
            \toprule
            \multirow{2}{*}{Statistics} & \multicolumn{2}{c}{SGAT4PASS} & \multicolumn{2}{c}{OmniREL-Reproj} \\
                                        & mIoU        & PAcc      & mIoU      & PAcc    \\ \midrule
            Mean                        & 55.984      &  82.887       &  65.040 ({\color{red}{+9.056}})    & 90.303 ({\color{red}{+7.416}}) \\
            Variance                    & 0.066       & 0.020         &  0.021 ({\color{red}{-0.045}})     & 0.003 ({\color{red}{-0.017}})  \\
            Range                       & 0.940       &0.478          &  0.544 ({\color{red}{-0.397}})     & 0.174 ({\color{red}{-0.304}})  \\ 
            \bottomrule
        \end{tabular}
    }
    \caption{
    Overall SGA validation performance comparison with SGAT4PASS~\cite{li2023sgat4pass} on Stanford2D3D Panoramic datasets. 
    ``PAcc'' means the pixel accuracy. 
    \ours earns significant mean performance and robustness improvement. 
    For fair comparison, the SGA validation setting and data augmentation in training is the same as SGAT4PASS~\cite{li2023sgat4pass}. 
    }
    \label{tab:SGAT}
\end{table}
\begin{table}[]
    \footnotesize
    \centering
    \resizebox{0.22\textwidth}{!}{
    \begin{tabular}{c|c}
    \hline
    D Info.       & mIoU  \\
    \hline
    RGB          & 51.25  \\
    RGB-D        & 59.03 \\
    RGB-ReD      & 62.02 \\
    RGB-EGVIA    & 63.35 \\ 
    RGB-Angle    & 59.76 \\
    RGB-Height   & 62.47 \\
    RGB-LOA      & 61.28 \\
    RGB-REL      & 64.47 \\
    \hline
    RGB-HHA-DyMM & 64.59 \\
    RGB-REL-DyMM & 66.73 \\
    \hline
    RGB-HHA-SMMF & 65.43 \\
    RGB-REL-SMMF & \textbf{67.37} \\
    \hline
    \end{tabular}
    }
    \caption{The effect of each module on Stanford2D3D Panoramic datasets fold 1 with traditional metrics. 
    The ``RGB-X'' means using both RGB and X channels. 
    ``-DyMM'' / ``-SMMF'' means using~\cite{xue2023dynamic}  / SMMF. 
    Note that Angle / Height is $\hat{A_p}$ and $\hat{H_p}$. 
    }
    \label{tab:ablation}
\end{table}


\section{Experiments}          
\subsection{Datasets and Protocols}  
We validate \ours on Stanford2D3D Panoramic datasets~\cite{armeni2017joint}. 
It has 1413 panoramic images with 13 labeled semantic classes. 
It has 3 official folds, fold 1, fold 2, and fold 3. 
We follow previous works~\cite{li2023sgat4pass, zhang2022bending, zhang2024behind} setting. 

Our experiments were conducted with a server with 8 NVIDIA GeForce RTX 3090 GPUs.
We use CMX~\cite{zhang2023cmx} as our baseline and the height / width of input images are 2048 / 4096 pixels, and set $\alpha / \lambda / m / n = 45^{\circ} / 0.5 / 3 / 7$.
Each region is $1080 \times 1080$, and the region stride is 720 pixels at inference.  
We overall follow~\cite{xue2023dynamic} for training and set soft / hard training epochs is both 100, and $T$ decays from 1 to 0.1 in the 100 soft training epochs.

\textbf{SGA validation}
Most PASS datasets follow a unified ERP way to deal with original $360^{\circ}$ data so that whether PASS models have the potential to overfit the ERP pattern is questionable.
It may have poor 3D robustness so that SGA validation~\cite{li2023sgat4pass} is also used in validation. 
In detail, ``Mean''  means the average of all results (e.g., mIoU, per pixel accuracy, etc.). 
For a general rotation in a 3D space, the angles of yaw / pitch / roll are $\alpha / \beta / \gamma$. 
Similar to~\cite{li2023sgat4pass},we set 3D rotation disturbance is at most $5^{\circ}$ / $5^{\circ}$ / $360^{\circ}$ of $\beta$ / $\gamma$ / $\alpha$. 
We set $n_{\alpha}=4$ ($0^{\circ}, 90^{\circ}, 180^{\circ}, 270^{\circ}$), $n_{\beta}=2$ ($0^{\circ}, 5^{\circ}$), and $n_{\gamma}=2$ ($0^{\circ}, 5^{\circ}$). 
``Variance''  means the variance of all results.  
``Range''  means the gap between the maximum and minimum results. 
Compared to traditional one, SGA validation avoids models gaining performance by fitting the datasets ERP pattern and reflects objective 3D robustness. 


\subsection{Performance Comparison}
\label{ssec:SOTA}
We first compare with several recent SOTA methods with traditional metrics. 
Also, we compare \ours with the baseline using HHA and SGAT4PASS~\cite{li2023sgat4pass} in detail with SGA metrics to validate 3D robustness. 

\textbf{Traditional metrics.}
Comparison results on Stanford2D3D Panoramic datasets with SOTA methods are shown in~\cref{tab:sota}. 
Following recent work, we report the performance of both official fold 1 and the average performance of all three official folds. 
A considerable improvement of 2.35\% / 3.39\% on the mIoU performance of 3 official folds on average / official fold 1 is observed and  our REL representation is more suitable for model training.
\begin{figure*}[tb]
    \centering
        \includegraphics[width=1\linewidth]{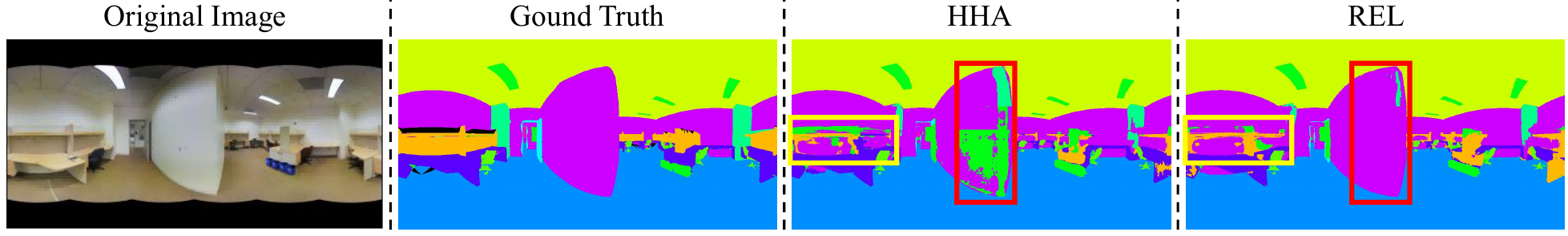}
        \caption{Comparison of HHA and REL results. 
        Especially the region with changeable LOA, REL reduces the noise on wall in the red box when the details are clearer in the yellow box.
        }
    \label{fig:visualization}
    \centering
\end{figure*}

\begin{figure}[tb]
    \centering
        \includegraphics[width=0.8\linewidth]{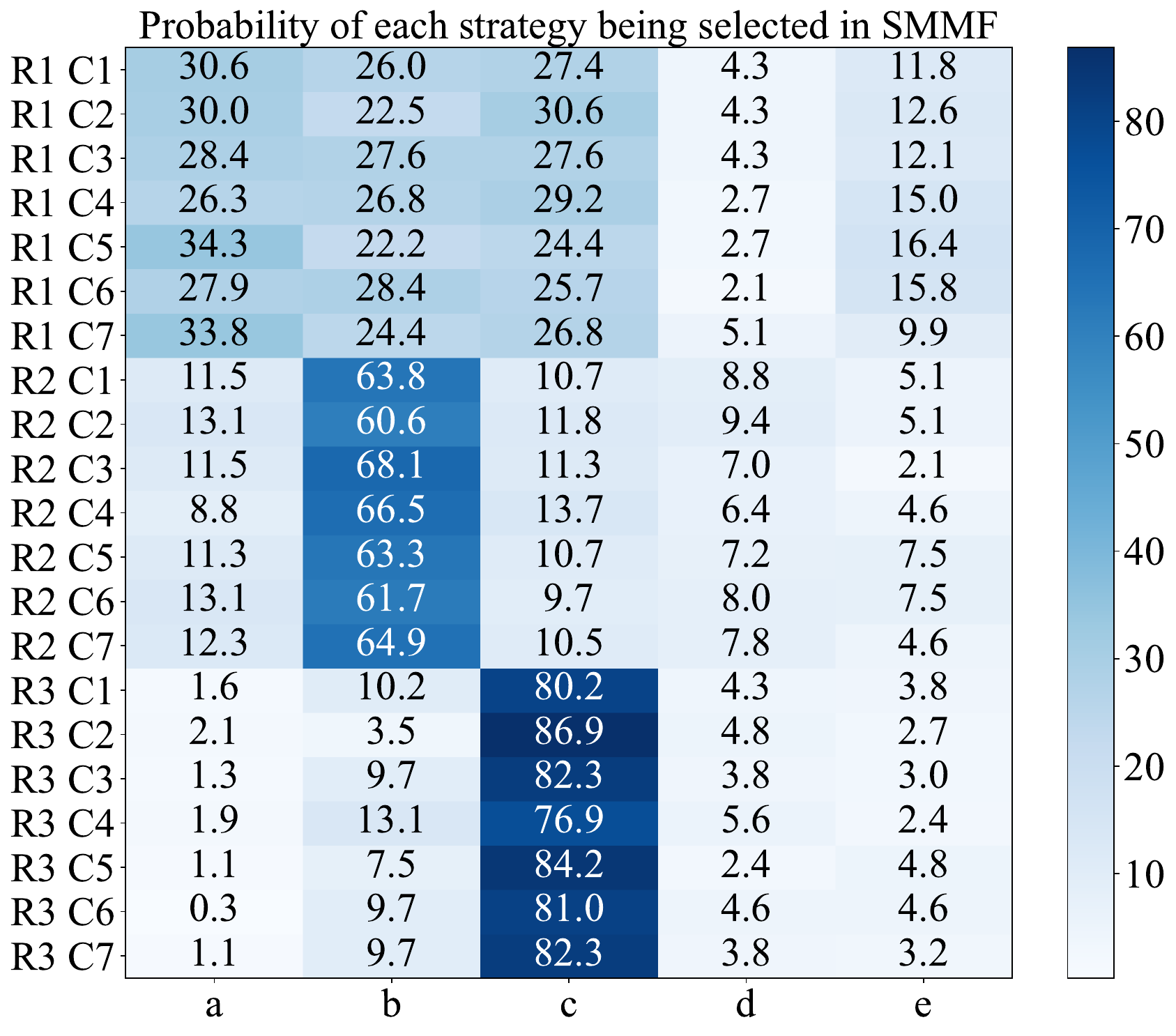}
        \caption{The visualization of the probability of each fusion strategy being selected for each region at inference. 
        ``R$X$ C$Y$'' means the region in $X^{th}$ row and $Y^{th}$ column. 
        The value is the probability (\%) of choosing this fusion strategy throughout the inference process. 
        a / b / c / d / e indicates $\mathbf{g} = [0,0,0,0]$ / $[1,0,0,0]$ / $[1,1,0,0]$ / $[1,1,1,0]$ / $[1,1,1,1]$. 
        }
    \label{fig:RAMM}
    \centering
\end{figure}

\textbf{SGA metrics.}
Comparison results with HHA on Stanford2D3D Panoramic datasets in SGA validation~\cite{li2023sgat4pass} are shown in~\cref{tab:STA_SGAM_BIG}. 
~\cref{tab:SGAMBIG} is the detailed performance of each situation. 
The models are the same as ``RGB-HHA-SMMF'' and ``RGB-REL-SMMF'' in~\cref{tab:ablation}. 
For mean mIoU / pixel accuracy, an improvement of more than 3.1\% / 1.1\% is achieved, respectively. 
Furthermore, our variance / fluctuation is about 30\% / 60\% of the HHA ones. 
These results show that \ourHHA have a much better robustness than HHA. 

Also, we compare the SGA validation~\cite{li2023sgat4pass} with SGAT4PASS having good resistance to 3D disturbances in~\cref{tab:SGAT}. 
For fair comparison, the SGA validation setting and data augmentation in training are the same as SGAT4PASS~\cite{li2023sgat4pass}.
Note that it is not the same as in~\Cref{tab:sota},~\Cref{tab:STA_SGAM_BIG}, and~\Cref{tab:SGAMBIG}, which is optimized for SGA metrics instead of traditional metrics.
For SGAT4PASS~\cite{li2023sgat4pass} is RGB-based method, we mainly focus on variance and fluctuation. 
Our variance / fluctuation is also about 30\% / 60\% of the SGAT4PASS ones. 
The detailed performance is shown in Section B ``The Detail Results Compared with SGAT4PASS'' in the supplementary material.

\subsection{Ablation Study}
\label{ssec:ablation}
We discuss the contribution of REL representation and SMMF, the contribution of ReD, EGVIA, and LOA in REL representation, and the contribution of height and \Eangle, in EGVIA. 

\textbf{Effect of REL Representation and SMMF}
The effectiveness of REL representation and SMMF is studied on Stanford2D3D Panoramic datasets official fold 1 with traditional metrics as shown in~\Cref{tab:ablation}. 
Using our REL representation can improve RGB-D mIoU by 5.44\%. 
Dynamic fusion benefit is considerable, and REL using DyMM gains 2.26\% improvement when using our SMMF can further improve 0.64\%. 
Similar performance improvements also appears on HHA.

\textbf{Effect of Each Channel in REL}
The effectiveness of all channels, ReD, EGVIA, and LOA, in REL representation are studied on Stanford2D3D Panoramic datasets official fold 1 with traditional metrics as shown in~\Cref{tab:ablation}. 
Using ReD / EGVIA / LOA channel alone improves the RGB-D mIoU by 2.99\% / 4.32\% / 2.25\%. 

\textbf{Effect of the different content of EGVIA}
The effectiveness of the height and \Eangle, in EGVIA channel is studied on Stanford2D3D Panoramic datasets official fold 1 with traditional metrics as shown in~\Cref{tab:ablation}. 
Using \Eangle / Height alone improves the RGB-D mIoU by 0.73\% / 3.44\%, when using a EGVIA improves the RGB-D mIoU by 4.32\%.

\subsection{Visualization}
\label{ssec:discussion}

\textbf{Visualization of different depth representation. }
We visualize the results of HHA and REL. 
Results show that REL reduces the noise and makes details clearer.

\textbf{SMMF Spherical consistency.} 
We visualize the strategy choice in SMMF to show whether \ours learns spherical geometry. 
It is easy to learn that due to different ERP patterns, the choice at the same latitude $\phi$ should be similar, or the different expansion for the cylinder side surface may lead to performance fluctuation. 
As shown in~\cref{fig:RAMM}, it is the fusion strategy selections for all 21 regions at inference. 
We can learn that the regions in the same row (latitude) have a similar fusion strategy choice, indicating the spherical consistency. 

More visualization are shown in Section C `More Visualizations'' in the supplementary material.


\section{Conclusion}
\ours is composed of a depth representation, \ourHHA, specifically for panoramic images and a region-level spherical-dynamic Multi-Modal Fusion to effectively integrate RGB and REL input. 
It applies spherical geometry prior to depth representation and fusion process, and shows considerable performance improvement and 3D robustness (e.g., resisting 3D disturbances) on popular datasets with both traditional and SGA metrics. 

{
    \small
    \bibliographystyle{ieeenat_fullname}
    \bibliography{main}
}


\end{document}